\documentclass[conference]{IEEEtran}
\IEEEoverridecommandlockouts
\usepackage[caption=false,font=footnotesize]{subfig}
\usepackage{cite}
\usepackage{amsmath,amssymb,amsfonts}
\usepackage{algorithmic}
\usepackage{graphicx}
\usepackage{textcomp}
\usepackage{xcolor}
\usepackage[]{algorithm2e}
\usepackage{graphicx}  
\usepackage{booktabs}       
\usepackage{amsmath}
\usepackage{makecell}
\usepackage{array,multirow}
\def\BibTeX{{\rm B\kern-.05em{\sc i\kern-.025em b}\kern-.08em
		T\kern-.1667em\lower.7ex\hbox{E}\kern-.125emX}}
\begin{document}

\title{Students are the Best Teacher: Exit-Ensemble Distillation with Multi-Exits}
	
\author{\IEEEauthorblockN{Hojung Lee}
\IEEEauthorblockA{\textit{School of Integrated Technology, Yonsei University, Korea} \\
hjlee92@yonsei.ac.kr}
\and
\IEEEauthorblockN{Jong-Seok Lee}
\IEEEauthorblockA{\textit{School of Integrated Technology, Yonsei University, Korea} \\
jong-seok.lee@yonsei.ac.kr}
}
	
\maketitle

\begin{abstract}
	This paper proposes a novel knowledge distillation-based learning method to improve the classification performance of convolutional neural networks (CNNs) without a pre-trained teacher network, called exit-ensemble distillation. 
	Our method exploits the multi-exit architecture that adds auxiliary classifiers (called exits) in the middle of a conventional CNN, through which early inference results can be obtained. 
	The idea of our method is to train the network using the ensemble of the exits as the distillation target, which greatly improves the classification performance of the overall network. 
	Our method suggests a new paradigm of knowledge distillation; unlike the conventional notion of distillation where teachers only teach students, we show that students can also help other students and even the teacher to learn better. 
	Experimental results demonstrate that our method achieves significant improvement of classification performance on various popular CNN architectures (VGG, ResNet, ResNeXt, WideResNet, etc.). 
	Furthermore, the proposed method can expedite the convergence of learning with improved stability.
	Our code will be available on github.
\end{abstract}

\section{Introduction}

Knowledge distillation (KD) \cite{Hinton14, Gou20} is a method to transfer learned information from a pre-trained large-sized network (teacher) to a small-sized one (student).
It was proposed as a way of model compression to reduce the size of the network while maintaining the performance.
In addition, it has been shown that KD can be also used to improve the performance.
An example is the self distillation method \cite{Furl18}, which uses the same network architecture for both the teacher and the student.
Besides, it has been shown that an ensemble of multiple trained models having the same structure but different weight parameters after training can be a teacher, which improves the performance of KD further \cite{Asif20}. 
However, a disadvantage of these KD methods is that they require pre-trained models, which involves additional computational complexity. 
When an ensemble model is used as a teacher, the burden of training teacher models becomes even more significant.

In this paper, we propose a novel KD method for performance improvement of convolutional neural networks (CNNs), called exit-ensemble distillation (EED), which eliminates the requirement of pre-trained teacher models.
The proposed method uses a multi-exit architecture where small-sized auxiliary classifiers called exits, each of which typically consists of a few convolutional layers and an output layer, are attached at certain intermediate layers of the main network \cite{Teer16}.
The key idea of our EED method is that the ensemble of all the classifiers in the network can serve as a good teacher for training the classifiers themselves to enhance the classification performance of the network.
Since each classifier (from the input to each exit or the final output) has a distinct structure, there exist diversity and complementarity in the outputs and features of the classifiers, which is the basis of the success of the ensemble approach.
The loss function to train the whole network consists of the conventional cross-entropy loss terms and the new distillation loss terms at all exits and the final output.

Previous distillation methods using multi-exit structures \cite{Phuong19, Beyour19, Li19} consider the exists as students and the main network as a teacher, and expect that distillation of the teacher's knowledge to the students eases learning of the network, especially for the early layers.
In this sense, these methods can be considered as one-way learning approaches.
On the other hand, our method is bidirectional in that each of the exits and the final output plays roles of both the teacher and the student.
Therefore, our method introduces a new paradigm of KD where small-sized networks (students in the conventional sense) help learning of large-sized networks (teachers in the conventional sense), which has not been explored previously to the best of our knowledge.

The proposed EED method have several advantages over existing approaches.
First, our method does not rely on specific model architectures but can be applied to generic CNNs by attaching exits at certain intermediate layers to create auxiliary classifiers.
Second, our method does not require a pre-trained model, since the whole network including the main network and the exits evolves concurrently via training.
Third, our method yields significant improvement of classification performance compared to previous approaches.

The contribution of this paper is summarized as follows.
\begin{itemize}
	\item We propose the EED method using multi-exit structures. Experimental results demonstrate that our method achieves better classification performance than previous distillation methods \cite{Teer16, Beyour19, Phuong19}.
	\item We show that our method makes the network learn class-specific features better. Moreover, our method enables fast and stable learning.
	\item We suggest a new possibility of KD to transfer knowledge between teachers and students in a bidirectional manner.
\end{itemize}

The rest of the paper is organized as follows.
Section \ref{sec:Related} provides a brief survey of the related work.
Section \ref{sec:method} explains the proposed EED method.
Extensive experimental results are provided in Section \ref{sec:exp}.
Finally, conclusion is given in Section \ref{sec:conclusion}.

\begin{figure*}[t]
	\begin{center}
		\includegraphics[width=0.9\linewidth]{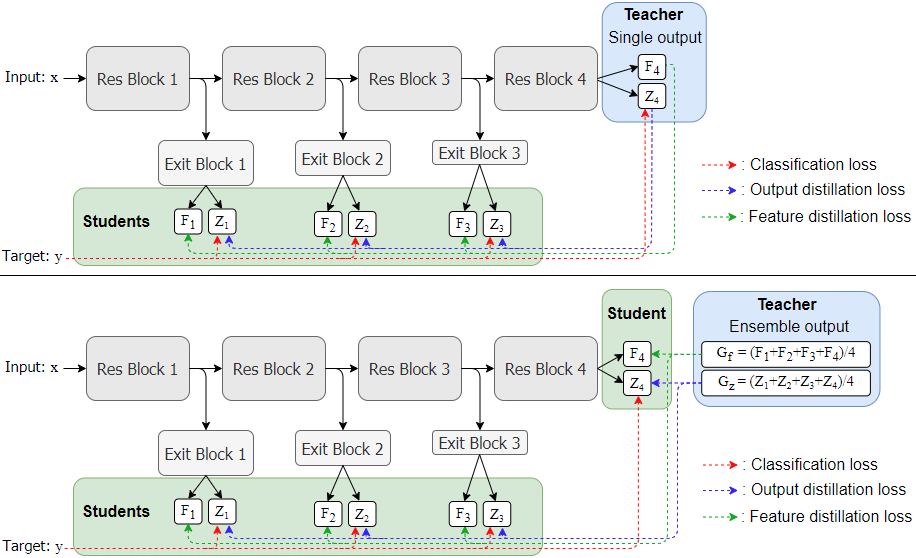}
	\end{center}
	\caption{Illustration of the self distillation \cite{Beyour19} (top) and the proposed EED (bottom). $F_i$ means the feature and $Z_i$ means the logits.}
	\label{fig:fig1}
\end{figure*}

\section{Related Work}
\label{sec:Related}

\subsection{Multi-Exit Architecture}
\label{sec:me}
In a typical CNN, a prediction result is obtained at the last layer through computation from the first layer to the last layer of the network.
To allow computational flexibility, a multi-exit architecture having multiple exits in the middle of the network can be used. 
Since each exit can produce a prediction result with a reduced amount of computation, anytime prediction can be performed by choosing a certain exit for prediction according to the computational budget.

The BranchyNet method \cite{Teer16} adds exits in a CNN and trains the network using the classification losses of all exits, which improves the classification performance by imposing a regularization effect and mitigating vanishing gradients. 
In \cite{Huang18, Yang20}, special multi-exit-based network structures are designed to enable efficient anytime prediction.

Our proposed method is based on multi-exit architectures derived from generic CNNs for versatility. 
In addition, we do not particularly aim at improving anytime prediction performance, but improving the prediction performance of the main network through distillation using the exits.

\subsection{Knowledge Distillation}
KD is a method using a pre-trained network as a teacher network to distill its learned knowledge to a smaller student network \cite{Hinton14}. 
A wide variety of KD techniques have been studied for various tasks, including classification model compression \cite{Li20}, object dectection \cite{Wang20}, pose estimation \cite{Zhang19}, and semantic segmentation \cite{Liu19}. 

In general, the teacher's knowledge to be distilled can be extracted at different levels, including logits \cite{Hinton14} and features \cite{Romero15}, which can be used directly or after transformation using another network or a kernel function \cite{Kim18, Heo19_1, Heo19}.
In addition, a teacher can be a single pre-trained network or an ensemble of multiple pre-trained networks. 

In a multi-exit architecture, the exits attached in the middle of the main network can be used as paths to deliver the information at the main network's output to early layers via distillation \cite{Phuong19, Beyour19, Li19}. 
In other words, the main network's output plays a role of the teacher and the exits are students for KD. 
Through this, each exit and shared parts of the main network can learn more general features, by not only following the true label but also receiving the knowledge of the final output that has the highest performance among all classifiers in the network. 
Consequently, the performance of not only the exits but also the final output is enhanced.

As a way to improve the multi-exit-based distillation further, we propose an ensemble method, where the teacher-student role assignment is not separate but the exits and the final output play roles of both the teacher and students at the same time. 
This greatly enhances the performance of the main network compared to the previous methods.

\subsection{Ensemble Approach}

The ensemble approach is popularly used for improved prediction performance. The outputs of the multiple models that have the same structure but are trained with different initializations are aggregated \cite{Simo15}. 
Thanks to the complementarity among the models \cite{Allen20}, this improves the test performance. However, the computational burden is significant to train multiple models.

An ensemble of multiple models can be used as a teacher for KD, i.e., the aggregated output of the ensemble can be transferred to a single student network \cite{Asif20}. 
However, this approach also has the issue of increased computational complexity.

A multi-exit architecture can be used to form an ensemble composed of the exits, where the models in the ensemble inherently have different structures. 
This alleviates the aforementioned computional burden for training because a multi-exit network has an only slightly increased model size compared to the corresponding network without exits. 
In the previous studies \cite{hojung19, Phuong19, Beyour19}, it has been shown that using the ensemble of exits during prediction can improve the classification performance compared to the single exits.

By taking one step further, we propose an ensemble method that uses each exit at the training stage for KD rather than at the inference stage.

\section{Proposed Method}
\label{sec:method}

In this section, we first describe the multi-exit (multi-classifier) architecture. Then, we present our proposed distillation method in detail, which is depicted in Figure \ref{fig:fig1}.

\subsection{Multi-Exit Architecture}

We build a multi-exit architecture by inserting $k-1$ exits into a generic CNN. 
The exits, denoted as $c_i$ ($i=1,...,k-1$), divide the CNN (main network) into $k$ blocks, which are denoted as $f_i$ ($i=1,...,k$).
Thus, the input data $x$ produces $k$ predictions (logits), $Z_i$ ($i=1,...,k$), by a single feedforward process, i.e.,
\begin{equation}
Z_{i} =
\begin{cases}
c_i(f_i(\cdots f_1(x))) & \text{if $i<k$}\\
f_i(\cdots f_1(x)) & \text{if $i=k$}
\end{cases}       
\end{equation}
In addition, the feature information right before the fully connected layer of each exit or the main network is denoted as $F_i$ ($i=1,...,k$).

For efficient operation and reduction of the training burden of a multi-exit architecture, the sizes of the exits are set small.
Therefore, the classifier for a later exit (from the input to the output of the exit) has a larger size and a larger learning capability, and consequently tends to show higher classification accuracy.
In addition, the outputs of the classifiers have diversity due to different structures and learning capabilities.

\subsection{Exit-Ensemble Distillation}

The basic way to train a multi-exit architecture is to use the joint classification loss $L_C$ with cross entropy (CE) for true label $y$ \cite{Teer16}, i.e.,
\begin{align}
\label{eq2}
L_C = \sum_{i=1}^{k} CE(q_i, y),
\end{align}
where $q_i = \frac{\text{exp}(Z_i/T)}{\sum_{j}^{} \text{exp}(Z_j/T)}$ is the softmax output and $T$ is a temperature.
Note that the added exits are learned only by the corresponding loss, but the shared parts of the main network among the exit classifiers are trained by the losses from multiple exits.
In addition to this classification loss, the distillation loss can be used as a training objective for exits to enhance the performance of the network \cite{Beyour19, Phuong19, Li19}, which uses the outputs and/or feature information of the main network (i.e., $Z_k$, $F_k$) as a teacher. 
Thus, the distillation loss $L_D$ is given by 
\begin{align}
\label{eq3}
L_{D} = \sum_{i=1}^{k-1} \bigg\{ \alpha l_z(Z_i, Z_k) + \beta l_f(F_i, F_k) \bigg\},
\end{align}
where $l_z$ and $l_f$ are the loss functions for the output distillation and the feature distillation, respectively, for which the Kullback-Leibler (KL) divergence\footnote{For KL, $Z_i$ and $Z_k$ are substituted with $q_i$ and $q_k$ in (\ref{eq3}), respectively.} or mean squared error (MSE) can be used. And, $\alpha$ and $\beta$ are coefficients.
Thus, the overall loss is given by
\begin{align}
\label{eq7}
L & = L_C + L_D \nonumber \\ 
& = \sum_{i=1}^{k} CE(q_i, y) + \sum_{i=1}^{k-1} \bigg\{ \alpha l_z(Z_i, Z_k) + \beta l_f(F_i, F_k) \bigg\}.
\end{align}

In using the distillation loss (\ref{eq3}), it is assumed that the outputs of the main network can act as a teacher for the exits because the performance of the main network is usually higher than that of the exits. 

However, we note that the ensemble of the exits and the main network can be an improved teacher thanks to their diversity and complementarity. Then, this superior ensemble knowledge can be transferred even to the main network's output as well as the exits.

The output ensemble signal and the feature ensemble signal are defined as
\begin{align}
\label{eq4}
G_z = \frac{1}{k} \sum_{i=1}^{k} Z_i, \\
G_f = \frac{1}{k} \sum_{i=1}^{k} F_i,
\end{align}
respectively.
Thus, our new distillation loss is given by
\begin{align}
\label{eq5}
L_{D} = \sum_{i=1}^{k} \bigg\{ \alpha l_z\left(Z_i, G_z\right) + \beta l_f\left(F_i, G_f\right) \bigg\}.
\end{align}
Finally, the overall loss for the multi-exit structure in the proposed EED method is written as
\begin{equation}
\label{eq6}
L = \sum_{i=1}^{k} \bigg\{ CE(q_i, y) + \alpha l_z\left(Z_i, G_z\right) + \beta l_f\left(F_i, G_f\right)  \bigg\}.
\end{equation}

The whole network including the main network and the attached exits is trained using this training loss from scratch.
All the exits and the main network output play roles of (parts of) the teacher by taking part in the ensemble.
At the same time, all of them play roles of students to learn the teacher's knowledge.
Therefore, the proposed method suggests a novel possibility of KD, which is completely different from the conventional way of assigning teachers and students separately.
Our method can be also called as a bidirectional KD approach in that each exit or the main network output delivers its knowledge as well as receives the ensemble knowledge.
\begin{table*}[t]
	\renewcommand{\arraystretch}{2.0}
	\begin{center}
		\begin{tabular}{|c|c|c|c|c|c|c|c|c|}
			\hline
			Network & Method & Training Loss & Exit 1 & Exit 2 & Exit 3 & Main \\\hline\hline
			\multirow{6}{*}{VGG19} 
			& Baseline & CE (without exits) & - & - & - & 73.27 \\ \cline{2-7}
			& \cite{Teer16} & CE (\ref{eq2}) & 68.29 & 71.46 & 74.81 & 74.60 \\ \cline{2-7}
			& \cite{Phuong19} & \makecell[l]{$\alpha=1 \: \text{(KL)},\: \beta=0$ in (\ref{eq7}) \\$\alpha=1 \: \text{(MSE)},\: \beta=0$ in (\ref{eq7})} & \makecell{68.59\\68.87} & \makecell{71.46\\71.67} & \makecell{75.02\\74.86} & \makecell{74.78\\74.87} \\ \cline{2-7}
			& \cite{Beyour19} & \makecell[l]{$\alpha=1 \: \text{(KL)},\: \beta=1$ in (\ref{eq7}) \\$\alpha=1 \: \text{(MSE)},\: \beta=1$ in (\ref{eq7})} & \makecell{68.67\\69.17} & \makecell{71.85\\71.52} & \makecell{74.75\\74.44} & \makecell{75.29\\74.26} \\ \cline{2-7}
			& EED & \makecell[l]{$\alpha=1 \: \text{(KL)}, \: \beta=0$ in (\ref{eq6})\\$\alpha=1 \: \text{(MSE)}, \: \beta=0$ in (\ref{eq6})\\$\alpha=1 \: \text{(KL)}, \: \beta=1$ in (\ref{eq6})\\$\alpha=1 \: \text{(MSE)}, \: \beta=1$ in (\ref{eq6})} & \makecell{68.80\\69.87\\69.38\\70.11} & \makecell{72.31\\72.81\\72.28\\72.70} & \makecell{75.89\\76.22\\75.68\\75.89} & \makecell{\textcolor{red}{75.98}\\\textcolor{red}{76.02}\\\textcolor{red}{75.69}\\\textcolor{red}{75.79}} \\
			\hline\hline
			\multirow{6}{*}{ResNet-18} 
			& Baseline & CE (without exits) & - & - & - & 77.71 \\ \cline{2-7}
			& \cite{Teer16} & CE (\ref{eq2}) & 70.70 & 74.95 & 78.00 & 78.58 \\ \cline{2-7}
			& \cite{Phuong19} & \makecell[l]{$\alpha=1 \: \text{(KL)},\: \beta=0$ in (\ref{eq7}) \\$\alpha=1 \: \text{(MSE)},\: \beta=0$ in (\ref{eq7})} & \makecell{71.96\\73.27} & \makecell{75.51\\77.17} & \makecell{77.72\\78.30} & \makecell{78.02\\78.31} \\ \cline{2-7}
			& \cite{Beyour19} & \makecell[l]{$\alpha=1 \: \text{(KL)},\: \beta=1$ in (\ref{eq7}) \\$\alpha=1 \: \text{(MSE)},\: \beta=1$ (\ref{eq7})} & \makecell{72.45\\73.38} & \makecell{75.31\\76.37} & \makecell{77.92\\78.03} & \makecell{78.29\\78.07} \\ \cline{2-7}
			& EED & \makecell[l]{$\alpha=1 \: \text{(KL)}, \: \beta=0$ in (\ref{eq6})\\$\alpha=1 \: \text{(MSE)}, \: \beta=0$ in (\ref{eq6})\\$\alpha=1 \: \text{(KL)}, \: \beta=1$ in (\ref{eq6})\\$\alpha=1 \: \text{(MSE)}, \: \beta=1$ in (\ref{eq6})} & \makecell{71.89\\73.26\\71.67\\73.58} & \makecell{75.81\\76.34\\75.39\\76.21} & \makecell{78.20\\78.75\\78.38\\78.59} & \makecell{\textcolor{red}{78.83}\\\textcolor{red}{79.25}\\\textcolor{red}{78.92}\\\textcolor{red}{79.24}} \\
			\hline\hline
			\multirow{6}{*}{ResNet-50} 
			& Baseline & CE (without exits) & - & - & - & 78.27 \\ \cline{2-7}
			& \cite{Teer16} & CE (\ref{eq2}) & 71.09 & 75.35 & 80.08 & 80.11 \\ \cline{2-7}
			& \cite{Phuong19} & \makecell[l]{$\alpha=1 \: \text{(KL)},\: \beta=0$ in (\ref{eq7}) \\$\alpha=1 \: \text{(MSE)},\: \beta=0$ in (\ref{eq7})} & \makecell{71.77\\73.25} & \makecell{76.92\\78.18} & \makecell{80.94\\80.24} & \makecell{80.99\\80.20} \\ \cline{2-7}
			& \cite{Beyour19} & \makecell[l]{$\alpha=1 \: \text{(KL)},\: \beta=1$ in (\ref{eq7}) \\$\alpha=1 \: \text{(MSE)},\: \beta=1$ (\ref{eq7})} & \makecell{71.14\\73.87} & \makecell{77.06\\78.81} & \makecell{80.58\\80.51} & \makecell{80.45\\80.23} \\ \cline{2-7}
			& EED & \makecell[l]{$\alpha=1 \: \text{(KL)}, \: \beta=0$ in (\ref{eq6})\\$\alpha=1 \: \text{(MSE)}, \: \beta=0$ in (\ref{eq6})\\$\alpha=1 \: \text{(KL)}, \: \beta=1$ in (\ref{eq6})\\$\alpha=1 \: \text{(MSE)}, \: \beta=1$ in (\ref{eq6})} & \makecell{72.28\\74.27\\71.56\\73.89} & \makecell{77.42\\78.51\\76.92\\78.66} & \makecell{81.28\\81.10\\81.34\\81.71} & \makecell{\textcolor{red}{81.34}\\\textcolor{red}{81.26}\\\textcolor{red}{81.35}\\\textcolor{red}{82.01}} \\
			\hline\hline
			\multirow{6}{*}{ResNet-101} 
			& Baseline & CE (without exits) & - & - & - & 79.15 \\ \cline{2-7}
			& \cite{Teer16} & CE (\ref{eq2}) & 71.96 & 76.61 & 81.35 & 81.14 \\ \cline{2-7}
			& \cite{Phuong19} & \makecell[l]{$\alpha=1 \: \text{(KL)},\: \beta=0$ in (\ref{eq7}) \\$\alpha=1 \: \text{(MSE)},\: \beta=0$ in (\ref{eq7})} & \makecell{72.01\\72.10} & \makecell{77.06\\78.23} & \makecell{81.19\\80.34} & \makecell{81.01\\80.18} \\ \cline{2-7}
			& \cite{Beyour19} & \makecell[l]{$\alpha=1 \: \text{(KL)},\: \beta=1$ in (\ref{eq7}) \\$\alpha=1 \: \text{(MSE)},\: \beta=1$ (\ref{eq7})} & \makecell{71.33\\72.56} & \makecell{76.70\\78.28} & \makecell{80.83\\80.29} & \makecell{80.94\\80.27} \\ \cline{2-7}
			& EED & \makecell[l]{$\alpha=1 \: \text{(KL)}, \: \beta=0$ in (\ref{eq6})\\$\alpha=1 \: \text{(MSE)}, \: \beta=0$ in (\ref{eq6})\\$\alpha=1 \: \text{(KL)}, \: \beta=1$ in (\ref{eq6})\\$\alpha=1 \: \text{(MSE)}, \: \beta=1$ in (\ref{eq6})} & \makecell{72.33\\73.34\\72.52\\73.45} & \makecell{77.64\\78.31\\78.02\\78.01} & \makecell{81.89\\81.15\\82.19\\81.40} & \makecell{\textcolor{red}{81.88}\\\textcolor{red}{81.19}\\\textcolor{red}{82.03}\\\textcolor{red}{81.24}} \\
			\hline
		\end{tabular}
	\end{center}
	\caption{Test accuracy (\%) of the network trained by each loss for CIFAR-100. The proposed distillation method achieves better performance of the main network than the previous methods in all cases, which are marked in red.}
	\label{tab:1}
\end{table*}

\begin{table*}[t]
	\renewcommand{\arraystretch}{1.5}
	\begin{center}
		\begin{tabular}{|c|c|c|c|c|c|c|c|c|c|}
			\hline
			Dataset & Network & Method & Exit 1 & Exit 2 & Exit 3 & Main \\\hline\hline
			\multirow{12}{*}{CIFAR-100} &
			\multirow{3}{*}{ResNet-152} & \cite{Teer16} & 71.05 & 78.60 & 81.87 & 81.80 \\ \cline{3-7} 
			& & \cite{Phuong19} & 72.03 & 79.85 & 81.99 & 82.07 \\ \cline{3-7} 
			& & EED & 71.62 & 78.67 & 82.07 & \textcolor{red}{82.29} \\ \cline{2-7} 
			& \multirow{3}{*}{ResNeXt29-4} & \cite{Teer16} & 68.71 & 77.87 & - & 79.62 \\ \cline{3-7} 
			& & \cite{Phuong19} & 70.77 & 78.84 & - & 79.75 \\ \cline{3-7} 
			& & EED & 68.99 & 78.12 & - & \textcolor{red}{80.14} \\ \cline{2-7} 
			& \multirow{3}{*}{WideResNet28-10} & \cite{Teer16} & 73.82 & 79.38 & - & 80.45 \\ \cline{3-7} 
			& & \cite{Phuong19} & 73.24 & 79.60 & - & 80.29 \\ \cline{3-7} 
			& & EED & 72.44 & 79.73 & - & \textcolor{red}{80.58} \\ \cline{2-7} 
			& \multirow{3}{*}{PyramidNet110-270} & \cite{Teer16} & 76.04 & 81.77 & - & 82.95 \\ \cline{3-7} 
			& & \cite{Phuong19} & 77.16 & 82.29 & - & 82.93 \\ \cline{3-7} 
			& & EED & 76.13 & 82.20 & - & \textcolor{red}{83.36} \\ \cline{2-7} 
			\hline\hline
			\multirow{6}{*}{ImageNet} 
			& \multirow{3}{*}{ResNet-18} & \cite{Teer16} & 47.71 & 55.49 & 62.44 & 66.74 \\ \cline{3-7} 
			& & \cite{Phuong19} & 48.15 & 55.71 & 62.59 & 66.66 \\ \cline{3-7} 
			& & EED & 48.54 & 55.89 & 62.36 & \textcolor{red}{66.94} \\ \cline{2-7} 
			& \multirow{3}{*}{ResNet-50} & \cite{Teer16} & 51.36 & 62.73 & 73.03 & 74.31 \\ \cline{3-7} 
			& & \cite{Phuong19} & 51.86 & 63.13 & 73.30 & 74.24 \\ \cline{3-7} 
			& & EED & 52.33 & 63.02 & 73.09 & \textcolor{red}{74.46} \\
			\hline
		\end{tabular}
	\end{center}
	\caption{Test accuracy (\%) of the network trained by each loss for CIFAR-100 and ImageNet. We use KL for \cite{Phuong19} and EED ($\beta=0$). The proposed EED method achieves better performance of the main network than the previous methods in all cases, which are marked in red.}
	\label{tab:2}
\end{table*}

\begin{table}[t]
	\renewcommand{\arraystretch}{1.5}
	\begin{center}
		\begin{tabular}{|l|c|c|c|c|c|c|c|}
			\hline
			Network & Exit 1 & Exit 2 & Exit 3 & Main \\
			\hline\hline
			VGG19 & 0.14 & 0.24 & 0.50 & 0.80 \\
			\hline
			ResNet-18 & 0.42 & 0.66 & 0.88 & 1.12 \\ 
			\hline
			ResNet-50 & 0.78 & 1.42 & 2.68 & 2.60 \\
			\hline
			ResNet-101 & 0.78 & 1.42 & 5.12 & 5.04 \\
			\hline
		\end{tabular}
	\end{center}
	\caption{Giga Flops required to obtain the output of each exit for CIFAR-100.}
	\label{tab:3}
\end{table}
\section{Experiments}
\label{sec:exp}
To demonstrate the performance of our method, we perform the image classification task with the CIFAR-100 \cite{Kri09} and ImageNet \cite{Deng09} datasets.

We use the ResNet \cite{he16deep} and other CNN architectures composed of residual blocks as main networks. 
They are mostly divided into three or four blocks by the pooling operation on the resolution of the intermediate feature map. 
Thus, we insert an exit network between the residual blocks.

To reduce the burden to train the added exits, the size of each exit should be set small. 
Thus, we use the minimum number of convolution layers, which can perform the necessary pooling operation through stride, and a fully connected layer. 
For example, we use three convolutional layers for the first exit, two for the second exit, and one for the third exit. 
In \cite{Beyour19}, the bottleneck structure is used for exits, but we experimentally found that the above structure performs better.

We use two loss functions for the output distillation loss, i.e., the KL loss and MSE loss. 
The KL loss is often used as an output distillation loss \cite{Beyour19, Phuong19}, but we find that there are many cases in which the MSE loss yields higher performance as shown in Table \ref{tab:1}. 
For the feature distillation, we use the MSE loss as in \cite{Heo19}.
We set both $\alpha$ and $\beta$ to 1 by default.

Other implementation details are given in the supplementary material.

\subsection{Performance Comparison}

To prove the effectiveness of our method, we compare the test performance of the method in \cite{Teer16} using (\ref{eq2}), the method in \cite{Phuong19} using (\ref{eq7}) with $\beta=0$, the method in \cite{Beyour19} using (\ref{eq7}) with $\beta=1$, and the proposed EED method using (\ref{eq6}), which is shown in Table \ref{tab:1}.
The table also shows the baseline performance without using the multi-exit structure.
First, when we compare the performance of the networks trained by only CE (the first two rows), the use of multi-exits allows us to obtain better performance of the main network in all cases.
The added losses from the exits give a regularization effect, which improves the test performance \cite{Teer16}.
Next, when distillation through the exits is applied using the output distillation \cite{Phuong19} ($\beta=0$), the performance of the early exits (exit 1 and exit 2) is particularly improved for all network structures with both KL and MSE.
Among them, ResNet-50 achieves the greatest performance improvement of the early exits by 2.16\% for exit 1 and 2.83\% for exit 2.
The advantage of the additional use of the feature distillation together with the output distillation \cite{Beyour19} ($\alpha=1, \: \beta=1$) is not clear.

Finally, it can be observed that the proposed method significantly improves the performance of the main network in all cases regardless of whether the feature distillation is used ($\alpha=1, \beta=1$) or not ($\alpha=1, \beta=0$).
The differences between the maximum performance of the main network without and with EED are 0.73\% for VGG19 (75.29\% $\rightarrow$ 76.02\%), 0.67\% for ResNet-18 (78.58\% $\rightarrow$ 79.25\%), 1.02\% for ResNet-50 (80.99\% $\rightarrow$ 82.01\%), and 0.89\% for ResNet-101 (81.14\% $\rightarrow$ 82.03\%).
In addition, the performance of all exits is also improved.
The differences between the maximum performance of exit 3 without and with EED are 1.20\% for VGG19 (75.02\% $\rightarrow$ 76.22\%), 0.45\% for ResNet-18 (78.30\% $\rightarrow$ 78.75\%), 0.77\% for ResNet-50 (80.94\% $\rightarrow$ 81.71\%), and 0.84\% for ResNet-101 (81.35\% $\rightarrow$ 82.19\%).

In addition, we evaluate our method on other popular CNN architectures including a larger ResNet (ResNet-152), ResNeXt \cite{Xie17}, WideResNet \cite{Zago16}, and PyramidNet \cite{Han17}. We also test our method on ImageNet for large-scale evaluation.
Since ResNext, WideResNet, and PyramidNet consist of three blocks, we use only two exits accordingly.
In Table \ref{tab:1}, the effectiveness of the feature distillation is not apparent, so it is excluded here. 
Moreover, it is hard to determine which is better between KL and MSE as the output distillation loss in Table \ref{tab:1}. 
Since KL is commonly used in the previous work, KL is employed in this experiment.

The results are shown in Table \ref{tab:2}. It can be seen that the proposed method improves the performance of the main network for the large-sized networks. Furthermore, our approach also achieves higher performance compared to the existing methods for ImageNet.
The performance improvement of the main network is 0.20\% for ResNet-18 and 0.15\% for ResNet-50.

As mentioned in Section \ref{sec:me}, the multi-exit structure is inherently capable of performing anytime prediction. 
Therefore, after training is done, adaptive inference can be performed according to the computation budget. 
In Table \ref{tab:3}, we show the number of floating-point operations (FLOPs) required to obtain the output from each exit or the main network. FLOPs for the outputs of early exits (exit 1 and exit 2) are relatively small compared to the main network.
For example, with ResNet-101, the difference between the maximum performance of exit 2 and the main network is 3.72\%, but exit 2 only needs 28.2\% (1.42/5.04) of FLOPs required for the main network.
Therefore, the trained network can be efficiently used in accordance with the computing environment, where the proposed method helps reducing the accuracy drop.
In the case of ResNet-50 and ResNet-101, exit 3 needs more FLOPs than the main network, but this is because we do not prioritize anytime prediction for designing exits, and it is not related to the performance of the main network.

\subsection{Ablation Study}

\begin{table}[t]
	\renewcommand{\arraystretch}{1.5}
	\begin{center}
		\begin{tabular}{|l|c|c|c|c|c|c|c|}
			\hline
			Teacher & Main \\ \hline\hline
			Main 		 			     & 78.31 \\ \hline
			Exit1 + Main 			  	 & 78.80 \\ \hline
			Exit3 + Main 				 & 78.25 \\ \hline
			Exit1 + Exit3 + Main    	 & 78.99 \\ \hline
			Exit1 + Exit2 + Exit3 		 & 78.47 \\ \hline 
			Exit1 + Exit2 + Exit3 + Main & 79.25 \\ \hline			
		\end{tabular}
	\end{center}
	\caption{Test accuracy (\%) of ResNet-18 for CIFAR-100 for varying compositions of the teacher.}
	\label{tab:4}
\end{table}

\begin{figure}[t]
	\begin{center}
		\includegraphics[width=0.8\linewidth]{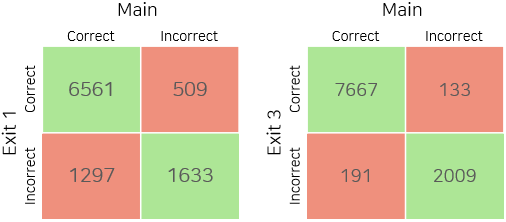}
	\end{center}
	\caption{Confusion matrices for the test data when ResNet-18 is trained using CE. The value in each cell indicates the number of test data classified correctly or incorrectly according to the classification results of the main network and an exit.}
	\label{fig:confusion}
\end{figure}

\begin{figure}[t]
	\centering
	\subfloat[]{%
		\includegraphics[width=0.38\textwidth]{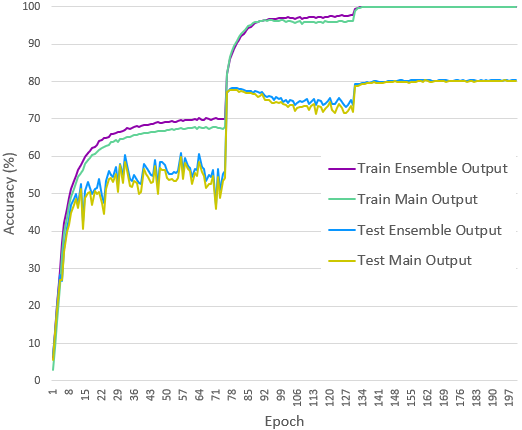}%
		\label{fig:_}
	}
	\hfill
	\subfloat[]{%
		\includegraphics[width=0.38\textwidth]{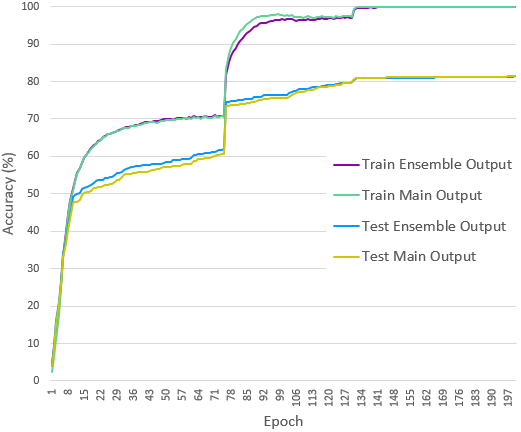}%
		\label{fig:__}
	}
	\caption{Accuracy curves for the training and test data of CIFAR-100 with ResNet-50 during training with (a) CE (b) EED.}
	\label{fig:2}
\end{figure}

\begin{figure*}[t]
	\begin{center}
		\includegraphics[width=0.85\linewidth]{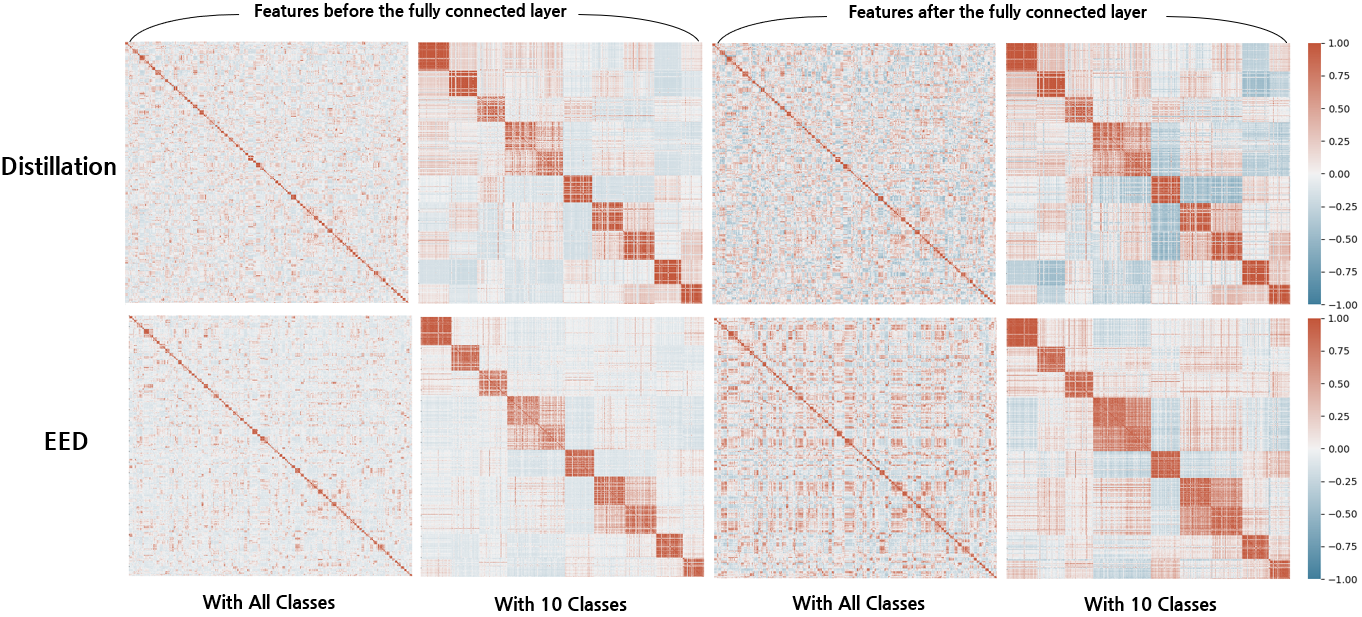}
	\end{center}
	\caption{Representation dissimilarity matrices of the layers of VGG19 trained by the conventional distillation method \cite{Phuong19} and the proposed EED method for label-ordered samples from CIFAR-100. A clear diagonal-block pattern indicates that clear inner-class representation has been learned.}
	\label{fig:RDM}
\end{figure*}

\begin{figure}[t]
	\begin{center}
		\includegraphics[width=0.8\linewidth]{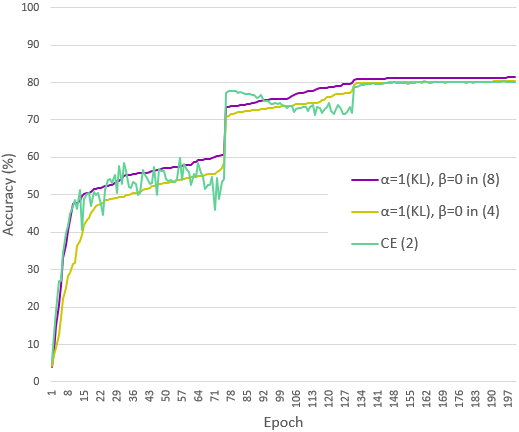}
	\end{center}
	\caption{Test accuracy curves of the main network during training for \cite{Teer16} using CE (\ref{eq2}), \cite{Phuong19} using (\ref{eq7}), and the proposed method using (\ref{eq6}).}
	\label{fig:4}
\end{figure}

We conduct an ablation study to investigate the performance of the main network by changing the composition of the exit-ensemble teacher in order to see how the composition of the ensemble target affects the performance of the network. 
The results are shown in Table \ref{tab:4}.
When the ensemble of exit 1 and the main network is used as a teacher, the performance of the main network is higher (78.80\%) than when the ensemble of exit 3 and the main network is used as a teacher (78.25\%).

Exit 1 has the smallest network size and the lowest classification performance, but has the least similarity with the main network in their outputs because it shares the least part with the main network.
Conversely, exit 3 has a large network size and high performance but has the most similarity with the main network because most parts are shared with the main network.
Thus, the ensemble of exit 1 and the main network, which involves high diversity, yields better performance than the ensemble of exit 3 and the main network.
To confirm this, we obtain the confusion matrix analyzing the classification results of the main network and each exit as shown in Figure \ref{fig:confusion}, when ResNet-18 is trained with CE. 
Indeed, the difference in prediction between exit 1 and the main network is much greater than the difference between exit 3 and the main network.

In addtion, the performance of the main network gradually improves as more exits are used together as the teacher.
If we use only the three exits without the main network, however, the performance is not improved a lot. 
Thus, the main network should be included as a part of the teacher to obtain high performance boost, implying that not only diversity but also quality of the teacher signals are crucial in our method.

\subsection{Analysis}
\label{sec:ana}

\subsubsection{Why does Exit-Ensemble Distillation Work?}

We assume that the performance improvement of the proposed EED method over the previous distillation methods is because of the power of the ensemble, i.e., the ensemble output achieves higher performance than the output of the main network itself.
To verify this assumption, we examine the performance of the main output and the ensemble output during training as shown in Figure \ref{fig:2}.
When we train the multi-exit structure using only CE, the accuracy of the ensemble output is always higher than that of the output of the main network. 
Therefore, our assumption that the ensemble output can be a better teacher than the output of the main is valid.
However, when the network is trained with EED, the accuracy difference between the ensemble output and the main network's output is reduced because the ensemble output is assigned to the main network as a distillation target and the main network mimics that.

\subsubsection{Exit-Ensemble Distillation Helps the Main Network Learn Class-Specific Representations.}

In order to analyze the trained networks, we obtain the representational dissimilarity matrices (RDMs) \cite{Kriege08} from the layers of VGG19 trained by the conventional distillation method \cite{Phuong19} and the proposed method. 
For each of 500 randomly chosen samples from CIFAR-100, the activations before and after the fully connected layer in the main network are recorded, and their correlation between two samples is measured, which is shown in Figure \ref{fig:RDM}. 
In the figure, clear diagonal-blocks indicate that for samples from the same class, the network generates highly correlated representations.
In order to see the results more clearly, the results for randomly chosen 10 classes are also displayed.

Overall, block-diagonal patterns become clearer when our EED method is used. 
Moreover, when EED is used, the correlation between different classes is closer to 0 than when conventional distillation is used, which appears to be more blurry in the off-diagonal region.
Thus, the proposed method helps the networks learn the image features that are not only common in the same class but also distinguished between different classes.

\subsubsection{Exit-Ensemble Distillation Achieves Stable Training.}

Another advantage of the proposed EED method is the stabilization of the learning process. 
As shown in Figure \ref{fig:4}, when learning is performed with only CE, significant irregular performance fluctuations are observed.
However, the accuracy curve is very stable when the distillation loss is added as in (\ref{eq7}). 
The reason is that when each exit uses only CE, learning proceeds according to its own loss regardless of the relationship with other exits. 
However, through the distillation loss, a common target is given, which increases the learning stability. 
Furthermore, when the proposed loss in (\ref{eq6}) is used, it is possible to greatly improve stability, convergence speed, and performance.

\section{Conclusion}
\label{sec:conclusion}
We proposed a novel distillation-based training method with auxiliary classifiers, called EED, which can improve the classification performance of CNNs. 
Our method suggests a way of creating an ensemble target so that the exits, which conventionally play only the role of students, can serve as a teacher at the same time, and applies this target to not only the students (exits) but also the teacher (the main network) for distillation.
We showed through the experiments that our method outperforms the previous methods.
We analyzed that the diversity and quality of the teacher signals in the ensemble target play an important role in the success of our method. 
Furthermore, we showed that our method helps the network learn class-specific feature representations. Finally, the additional advantage of stable training by our method was demonstrated.  

Since our method is network structure-independent, it can be applied to a wide variety of structures and can be also applied with other regularization and augmentation techniques, which can be investigated in the future.

\bibliographystyle{plain}
\bibliography{egbib}
	
\end{document}